\documentclass[a4paper,conference]{IEEEtran}
\ifCLASSINFOpdf
  \usepackage[pdftex]{graphicx}
\else
  \usepackage[dvips]{graphicx}
\fi

\usepackage{amsmath}
\usepackage{wasysym}

\usepackage{url}

\begin{document}
\title{Teacher-Student Training and Triplet Loss for Facial Expression Recognition under Occlusion}

\author{\IEEEauthorblockN{Mariana-Iuliana Georgescu\IEEEauthorrefmark{1}$^,$\IEEEauthorrefmark{2},
Radu Tudor Ionescu\IEEEauthorrefmark{2}}
\IEEEauthorblockA{
\IEEEauthorrefmark{1}Novustech Services, 12B Aleea Ilioara, Bucharest, Romania\\
Email: georgescu\_lily@yahoo.com
}
\IEEEauthorblockA{
\IEEEauthorrefmark{2}University of Bucharest, 14 Academiei, Bucharest, Romania\\
Email: raducu.ionescu@gmail.com}}


\maketitle

\begin{abstract}
In this paper, we study the task of facial expression recognition under strong occlusion. We are particularly interested in cases where 50\% of the face is occluded, e.g. when the subject wears a Virtual Reality (VR) headset. While previous studies show that pre-training convolutional neural networks (CNNs) on fully-visible (non-occluded) faces improves the accuracy, we propose to employ knowledge distillation to achieve further improvements. First of all, we employ the classic teacher-student training strategy, in which the teacher is a CNN trained on fully-visible faces and the student is a CNN trained on occluded faces. Second of all, we propose a new approach for knowledge distillation based on triplet loss. During training, the goal is to reduce the distance between an anchor embedding, produced by a student CNN that takes occluded faces as input, and a positive embedding (from the same class as the anchor), produced by a teacher CNN trained on fully-visible faces, so that it becomes smaller than the distance between the anchor and a negative embedding (from a different class than the anchor), produced by the student CNN. Third of all, we propose to combine the distilled embeddings obtained through the classic teacher-student strategy and our novel teacher-student strategy based on triplet loss into a single embedding vector.

We conduct experiments on two benchmarks, FER+ and AffectNet, with two CNN architectures, VGG-f and VGG-face, showing that knowledge distillation can bring significant improvements over the state-of-the-art methods designed for occluded faces in the VR setting. Furthermore, we obtain accuracy rates that are quite close to the state-of-the-art models that take as input fully-visible faces. For example, on the FER+ data set, our VGG-face based on concatenated distilled embeddings attains an accuracy rate of 82.75\% on lower-half-visible faces, which is only 2.24\% below the accuracy rate of a state-of-the-art VGG-13 that is evaluated on fully-visible faces. Given that our model sees only the lower-half of the face, we consider this to be a remarkable achievement. In conclusion, we consider that our distilled CNN models can provide useful feedback for the task of recognizing the facial expressions of a person wearing a VR headset.
\end{abstract}


% \fi
\IEEEpeerreviewmaketitle

\section{Introduction}

Facial expression recognition is an actively studied topic in computer vision, having many practical applications in various domains, such as detection of mental disorders, human behavior understanding and human-computer interaction. In this paper, we focus on the application of facial expression recognition to human-computer interaction in a virtual reality (VR) environment. More specifically, our aim is to design a system able to recognize the facial expressions of a user wearing a VR headset in order to automatically control and adjust the VR environment according to the user's emotions. To our knowledge, there are only a few approaches in this direction \cite{Georgescu-ICONIP-2019,Hickson-WACV-2019}, in which the main challenge is to deal with severe occlusions of the face caused by the fact that the user is wearing a VR headset covering the entire upper part of the user's face. Hickson et al.~\cite{Hickson-WACV-2019} proposed an approach that analyzes expressions from the eyes region captured with a camera mounted inside the VR headset, while Georgescu et al.~\cite{Georgescu-ICONIP-2019} proposed an approach that analyzes the lower-half of the face captured with a standard camera. We place ourselves in the same setting as Georgescu et al.~\cite{Georgescu-ICONIP-2019}, studying the task of facial expression recognition when the upper-half of the face is occluded. Different from Georgescu et al.~\cite{Georgescu-ICONIP-2019}, we propose to employ knowledge distillation to obtain more accurate convolutional neural networks (CNNs). We study two knowledge distillation approaches in order to distill information from CNNs trained on fully-visible faces to CNNs trained on occluded faces. First of all, we employ the classic teacher-student training strategy~\cite{Ba-NIPS-2014,Hinton-DLRL-2015}. Second of all, we propose a new approach for knowledge distillation based on triplet loss~\cite{Schroff-CVPR-2015}. During training, the goal is to reduce the distance between an anchor embedding, produced by a student CNN that takes occluded faces as input, and a positive embedding (from the same class as the anchor), produced by a teacher CNN trained on fully-visible faces, so that it becomes smaller than the distance between the anchor and a negative embedding (from a different class than the anchor), produced by the student CNN. To our knowledge, we are the first to apply triplet loss in order to distill knowledge into neural networks. Last but not least, we propose to combine the distilled face embeddings obtained through the classic teacher-student strategy and our novel teacher-student strategy based on triplet loss into a single face embedding vector, further boosting the performance gains.

We conduct experiments on two benchmark data sets, FER+~\cite{Barsoum-ICMI-2016} and AffectNet~\cite{Mollahosseini-TAC-2019}, comparing our models based on knowledge distillation with closely-related models~\cite{Georgescu-ICONIP-2019,Hickson-WACV-2019} designed for the VR setting, as well as with state-of-the-art models~\cite{Barsoum-ICMI-2016,Mollahosseini-TAC-2019,Guo-Sensors-2020,Ionescu-WREPL-2013,Kollias-BMVC-2019} that work on fully-visible faces. We note that the latter comparison with works such as~\cite{Barsoum-ICMI-2016,Mollahosseini-TAC-2019,Guo-Sensors-2020,Ionescu-WREPL-2013,Kollias-BMVC-2019} is not entirely fair, as these state-of-the-art models get to see the entire (non-occluded) faces at test time. This comparison is rather intended to provide some upper-bounds to the results that could be obtained on occluded faces. We also note that FER+ and AffectNet do not contain occluded faces, so need to modify the original images by blacking out regions to simulate the occlusions caused by the VR headset.

In all the experiments, the empirical results show that our distilled CNN models obtain superior results compared to the other models~\cite{Georgescu-ICONIP-2019,Hickson-WACV-2019} evaluated on occluded faces. Furthermore, the accuracy rates of our best models tested on lower-half-visible faces are between $2\%$ and $8\%$ under accuracy rates of the state-of-the-art CNN models, which are tested on fully-visible faces. We consider noteworthy the fact that the gap between our distilled models (evaluated on occluded faces) and the state-of-the-art ones (evaluated on fully-visible faces) is so small.

In summary, our contribution is threefold:
\begin{itemize}
\item We propose a novel knowledge distillation method based on triplet loss.
\item We propose to combine the classic teacher-student strategy with our strategy based on triplet loss using a late fusion strategy, i.e. by concatenating the distilled face embeddings.
\item We conduct experiments on two benchmarks, showing that knowledge distillation brings significant performance improvements in facial expression recognition under strong occlusion.
\end{itemize}

We organize the rest of this paper as follows. We discuss related work in Section~\ref{sec_related_work}. We present our knowledge distillation approaches in Section~\ref{sec_method}. We describe the empirical results in Section~\ref{sec_experiments}. Finally, we draw our conclusions in Section~\ref{sec_conclusion}.

\section{Related Work}
\label{sec_related_work}

\subsection{Facial Expression Recognition}

In the past few years, most works on facial expression recognition have focused on building and training deep neural networks in order to obtain state-of-the-art results~\cite{Barsoum-ICMI-2016,Kollias-BMVC-2019,Ding-FG-2017,Georgescu-Access-2019,Giannopoulos-2018,Hasani-CVPRW-2017,Hosseini-FG-2019,Hua-Access-2019,HU-ACII-2019,Kim-JMUI-2016,Li-MTA-2017,Li-CVPR-2017,Liu-CVPRW-2017,Li-ICPR-2018,Meng-FG-2017,Mollahosseini-CVPRW-2016,Tang-WREPL-2013,Wen-CC-2017,Wikanningrum-CENIM-2019,Yu-ICMI-2015,Zeng-ECCV-2018}. Engineered models based on handcrafted features~\cite{Ionescu-WREPL-2013, Al-Chanti-VISIGRAPP-2018,Shah-PRL-2017,Shao-PRL-2015} have drawn very little attention, since such models usually yield less accurate results compared to deep learning models. In~\cite{Barsoum-ICMI-2016, Guo-Sensors-2020}, the authors adopted VGG-like architectures. Barsoum et al.~\cite{Barsoum-ICMI-2016} designed a convolutional neural network specifically for the FER+ data set, consisting of $13$ layers (VGG-13). Guo et al.~\cite{Guo-Sensors-2020} focused on detecting the emotion on mobile devices, proposing a light-weight VGG architecture. In order to gain computational performance, they reduced the input size, the number of filters and the number of layers, and replaced the fully-connected layers with global average pooling. Their network consists of $12$ layers organized into $6$ blocks.

We note that most works studied facial expression recognition from static images, but there are also some works designed for video~\cite{Hasani-CVPRW-2017,Kaya-IVC-2017}. Hasani et al.~\cite{Hasani-CVPRW-2017} proposed a network architecture that consists of 3D convolutional layers followed by a Long Short-Term Memory (LSTM) network, extracting the spatial relations within facial images and the temporal relations between different frames in the video. 

Unlike other approaches, Meng et al.~\cite{Meng-FG-2017} and Liu et al.~\cite{Liu-CVPRW-2017} presented identity-aware facial expression recognition models. Meng et al.~\cite{Meng-FG-2017} proposed to jointly estimate expression and identity features through a neural architecture composed of two identical CNN streams, in order to alleviate inter-subject variations introduced by personal attributes and to achieve better facial expression recognition performance. Liu et al.~\cite{Liu-CVPRW-2017} employed deep metric learning and jointly optimized a deep metric loss and the softmax loss. They obtained an identity-invariant model by using an identity-aware hard-negative mining and online positive mining scheme. Li et al.~\cite{Li-CVPR-2017} trained a CNN model using a modified back-propagation algorithm which creates a locality preserving loss aiming to pull the neighboring faces of the same class together. Zeng et al.~\cite{Zeng-ECCV-2018} proposed a model that addresses the labeling inconsistencies across data sets. In their framework, images are tagged with multiple (pseudo) labels either provided by human annotators or predicted by learned models. Then, a facial expression recognition model is trained to fit the latent ground-truth from the inconsistent pseudo-labels. Hua et al.~\cite{Hua-Access-2019} proposed a deep learning algorithm consisting of three sub-networks of different depths. Each sub-network is based on an independently-trained CNN.

Different from all the works mentioned so far and many others~\cite{Barsoum-ICMI-2016,Ionescu-WREPL-2013,Kollias-BMVC-2019,Ding-FG-2017,Georgescu-Access-2019,Giannopoulos-2018,Hasani-CVPRW-2017,Hosseini-FG-2019,Hua-Access-2019,HU-ACII-2019,Kim-JMUI-2016,Li-MTA-2017,Li-CVPR-2017,Liu-CVPRW-2017,Li-ICPR-2018,Meng-FG-2017,Mollahosseini-CVPRW-2016,Tang-WREPL-2013,Wen-CC-2017,Wikanningrum-CENIM-2019,Yu-ICMI-2015,Zeng-ECCV-2018,Al-Chanti-VISIGRAPP-2018,Shah-PRL-2017,Shao-PRL-2015}, that recognize facial expressions from fully-visible faces, we focus on recognizing the emotion using only the lower part of the image. The number of works that focus on facial expression recognition under occlusion is considerably smaller~\cite{Georgescu-ICONIP-2019,Hickson-WACV-2019,Li-ICPR-2018}. Li et al.~\cite{Li-ICPR-2018} trained and tested a model on synthetically occluded images. They proposed an end-to-end trainable Patch-Gated CNN that automatically detects the occluded regions and focuses on the most discriminative non-occluded regions. Different from Li et al.~\cite{Li-ICPR-2018}, we consider a more difficult setting in which half of the face is completely occluded. In order to learn effectively in this difficult setting, we transfer knowledge from teacher networks that are trained on fully-visible (non-occluded) faces. 
 
Closer to our approach are the works designed for the difficult VR setting~\cite{Georgescu-ICONIP-2019,Hickson-WACV-2019}, in which a VR headset covers the upper side of the face. Hickson et al.~\cite{Hickson-WACV-2019} proposed a method that analyzes expressions from the eyes region. The eyes region is captured by an infrared camera mounted inside the VR headset, making the method less generic. Georgescu et al.~\cite{Georgescu-ICONIP-2019} proposed an approach that analyzes the mouth region captured with a standard camera. In this work, we use the same setting as Georgescu et al.~\cite{Georgescu-ICONIP-2019}, studying the task of facial expression recognition when the upper-half of the face is occluded. Different from Georgescu et al.~\cite{Georgescu-ICONIP-2019}, we propose to employ knowledge distillation to obtain more accurate CNNs. We study two knowledge distillation approaches in order to distill information from CNNs trained on fully-visible faces to CNNs trained on occluded faces. To our knowledge, we are the first to apply knowledge distillation in the context of facial expression recognition under strong occlusion.

\subsection{Knowledge Distillation}

Knowledge distillation~\cite{Ba-NIPS-2014,Hinton-DLRL-2015} is a recently studied approach~\cite{Feng-arxiv-2019,Lopez-ICLR-2016,Park-CVPR-2019,Yim-CVPR-2017,You-KDD-2017,Yu-CVPR-2019} that enables the transfer of knowledge between neural networks. Knowledge distillation is a framework that unifies model compression~\cite{Ba-NIPS-2014,Hinton-DLRL-2015,Feng-arxiv-2019} and learning under privileged information~\cite{Lopez-ICLR-2016,vapnik-vashist-nn-2009}, the former one being more popular than the latter. In model compression, knowledge from a large neural network~\cite{Ba-NIPS-2014,Park-CVPR-2019} or an ensemble of large neural models~\cite{Hinton-DLRL-2015,You-KDD-2017,Yu-CVPR-2019} is distilled into a small neural network, that runs efficiently during inference. In learning under privileged information, knowledge from a neural model training on privileged information (additional data representation not available at test time) is transferred to another neural model that does not have access to the privileged information. In our paper, we are not interested in compressing neural models, but in learning under privileged information. In particular, we study teacher-student training strategies, in which the teacher neural network can learn from fully-visible faces and the student neural network can learn from occluded faces only. In this context, hidden (occluded) face regions represent the privileged information.

To our knowledge, we are the first to propose the distillation of knowledge using triplet loss. We note that there are previous works~\cite{Feng-arxiv-2019,Park-CVPR-2019,You-KDD-2017,Yu-CVPR-2019} that distilled triplets or the metric space from a teacher network to a student network. Different from these methods, we do not aim to transfer the metric space learned by a teacher network, but to transfer knowledge from the teacher using metric learning, which is fundamentally different.

\section{Methods}
\label{sec_method}

To demonstrate that our knowledge distillation methods generalize across neural architectures, we employ two CNN models, namely VGG-f~\cite{Chatfield-BMVC-14} and VGG-face~\cite{Parkhi-BMVC-2015}. We opted for these particular models to allow a direct and fair comparison with Georgescu et al.~\cite{Georgescu-ICONIP-2019}. We note that VGG-f is pre-trained on object recognition, while VGG-face is pre-trained on face recognition. To obtain the teacher networks, we fine-tune VGG-f and VGG-face on facial expression recognition from fully-visible faces. Similarly, we obtain the student networks by fine-tuning VGG-f and VGG-face on facial expression recognition from lower-half-visible faces. In order to fine-tune the models, we employ the same hyperparameters as Georgescu et al.~\cite{Georgescu-ICONIP-2019}. Unlike Georgescu et al.~\cite{Georgescu-ICONIP-2019}, we next employ one of two teacher-student training strategies to learn privileged information from a teacher CNN to a student CNN. We note that the two architectures, VGG-f and VGG-face, are never mixed in the training process. In other words, we use the teacher VGG-f to distill knowledge into the student VGG-f, and independently, the teacher VGG-face to distill knowledge into the student VGG-face. Our goal is not to compress the models, but to improve performance using the privileged information accessible to the teacher networks. In our case, the privileged information is represented by the upper-half of the face, which is occluded (not available) at test time. 

We next describe in detail the teacher-student strategies for facial expression recognition under strong occlusion.

\begin{figure*}[!t]
\centering
\includegraphics[width=0.6\linewidth]{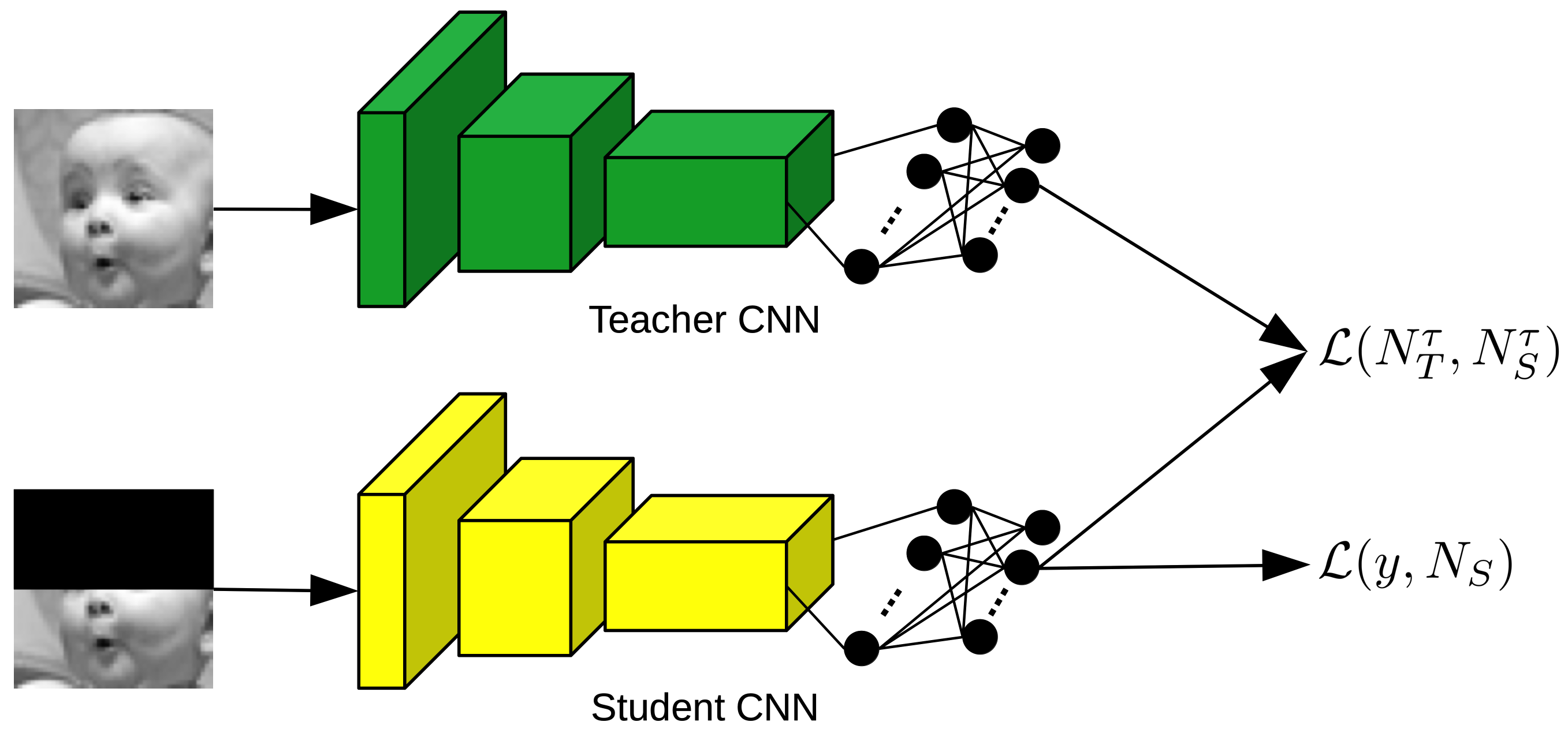}
%\vspace{-0.3cm}
\caption{The standard teacher-student training pipeline for facial expression recognition on severely occluded faces. The teacher CNN takes as input non-occluded (fully-visible) faces, having access to privileged information. The student CNN takes as input only occluded (lower-half-visible) faces, but learns useful information from the teacher CNN model. The loss functions $\mathcal{L}(y, N_S)$ and $\mathcal{L} (N_{T}^\tau , N_S^{\tau})$ are defined in Equation~\eqref{eq_loss_KD}. Best viewed in color.}\label{fig_teacher_student}
%\vspace{-0.3cm} 
%\vspace{-0.3cm}
\end{figure*}

%\vspace{-0.1cm}
\subsection{Standard Teacher-Student Training}
\label{sec_train_teacher}
%\vspace{-0.1cm}

Ba et al.~\cite{Ba-NIPS-2014} discussed the idea of model compression in the context of deep learning. 
Model compression refers to training a compact (shallow) model to approximate the function learned by a more complex (deeper) model. Hinton et al.~\cite{Hinton-DLRL-2015} further developed the idea and proposed to distill the knowledge from an ensemble of models into a single neural network, to achieve faster inference time and improved performance. Hinton et al.~\cite{Hinton-DLRL-2015} suggested that knowledge can be transferred in a straightforward manner by training the distilled (student) network using a soft target distribution that is produced by the ensemble (teacher) model. Improving this approach is possible if the correct labels for the chosen training samples are also known by the student network. In this case, we can employ a weighted average of two different loss functions, thus obtaining a single knowledge distillation (KD) loss function:
\begin{equation}\label{eq_loss_KD}
\mathcal{L}_{KD'} (\theta_S) = (1 - \lambda) \mathcal{L}(y, N_S) + \lambda \mathcal{L} (N_{T}^\tau, N_S^{\tau}),
\end{equation}
where $\theta_S$ are the weights of the student network $S$, $y$ are the target labels, $N_T$ and $N_S$ are the outputs of the teacher network $T$ and the student network $S$, respectively, $N_{T}^\tau$ and $N_S^{\tau}$ are the softened outputs of $T$ and $S$, respectively, and $\tau > 1$ is a temperature parameter for the softening operation. We note that $N_{T}^\tau$ and $N_S^{\tau}$ are derived from the pre-softmax activations $A_T$ and $A_S$ of the teacher network and the student network, respectively:
\begin{equation}
N_{T}^\tau = softmax \left( \frac{A_T}{\tau} \right), N_{S}^\tau = softmax \left( \frac{A_S}{\tau} \right).
\end{equation}
In Equation~\eqref{eq_loss_KD}, the first loss function $\mathcal{L}(y, N_S)$ is the cross-entropy with respect to the correct labels. The second loss function $\mathcal{L} (N_{T}^\tau, N_S^{\tau})$ is the cross-entropy with respect to the soft targets provided by the teacher network. Hinton et al.~\cite{Hinton-DLRL-2015} suggested that the second objective function must have a higher weight with respect to the first objective function in Equation~\eqref{eq_loss_KD}. Therefore, we set $\lambda = 0.7$ for both VGG-f and VGG-face. We also set the same temperature ($\tau = 20$) for both networks.

Lopez-Paz et al.~\cite{Lopez-ICLR-2016} proposed a generalized knowledge distillation paradigm that allows to learn not only from a different (teacher) network, but also by using a different data representation. Indeed, the teacher-student paradigm presented in Equation~\eqref{eq_loss_KD} can also be applied to learn privileged information~\cite{vapnik-vashist-nn-2009} that is available only to the teacher network. This approach is suitable in our setting that involves learning from severely occluded faces, as we aim to learn from a teacher network that has access to privileged information, i.e. non-occluded faces. We therefore choose as teacher networks the CNN models that are trained on fullly-visible face images. We stress out that VGG-f and VGG-face are used both as teacher networks and as student networks, but without mixing the architectures, i.e. the teacher and the student networks always have the same architecture. The fact that the teacher's architecture is not deeper (as commonly agreed in~\cite{Ba-NIPS-2014,Hinton-DLRL-2015}) is less important in our context. What matters most is that the teacher network is trained on non-occluded faces and its soft labels provide additional information that is not seen by the student network in the input images, since the faces are occluded. Our first teacher-student framework is illustrated in Figure~\ref{fig_teacher_student}.

\begin{figure*}[!t]
\centering
\includegraphics[width=0.92\linewidth]{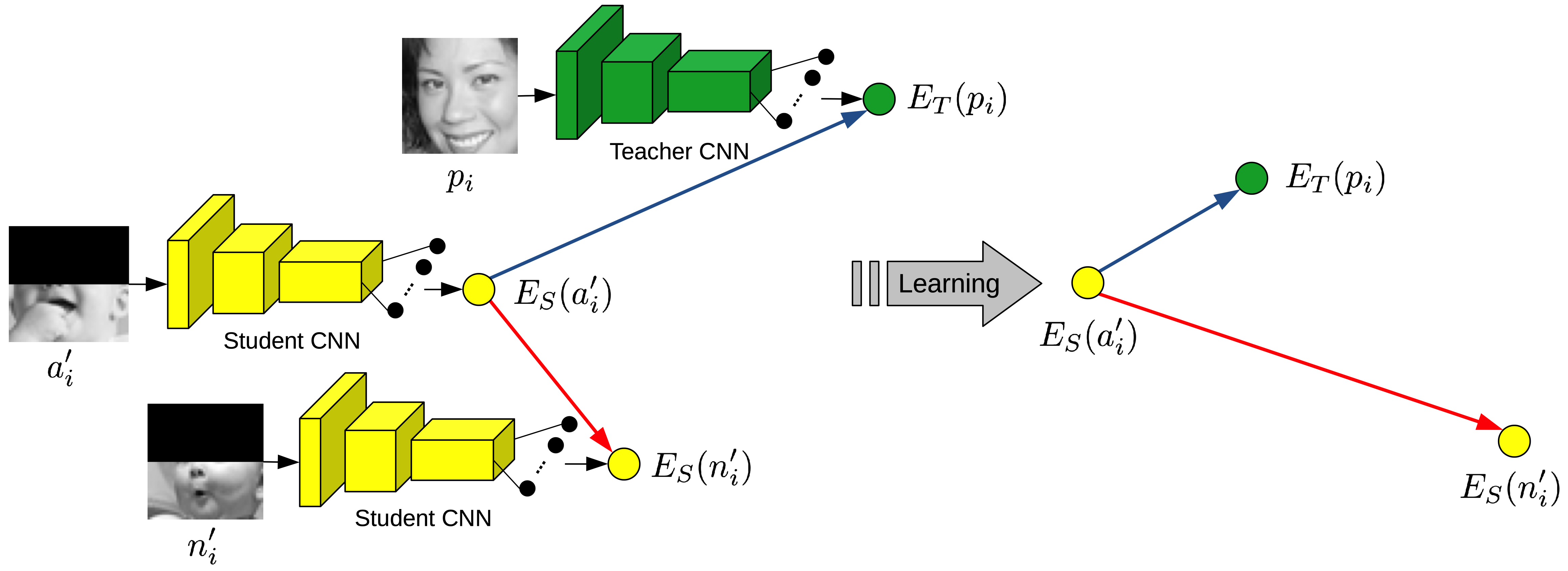}
%\vspace{-0.3cm}
\caption{The teacher-student training based on triplet loss for facial expression recognition on severely occluded faces. During training, we modify the weights of the student network such that the distance $\Delta(E_S(a'_i), E_T(p_i))$ becomes smaller than the distance $\Delta(E_S(a_i'), E_S(n'_i))$. Best viewed in color.}\label{fig_triplet_loss}
%\vspace{-0.3cm}
\end{figure*}

\subsection{Teacher-Student Training with Triplet Loss}
\label{sec_train_teacher_triplet}

In standard teacher-student training, the goal is to train the student network to reproduce the output class probabilities of the teacher network, in addition to predicting the correct labels. We believe that the same effect can be achieved by training the student network to reproduce the face embeddings (the activations from the penultimate layer) of the teacher network. However, we do not necessarily want to obtain similar embeddings when the student network takes an input image from a different class than the teacher network. In other words, it is sufficient to ensure that the embeddings produced by the student and the teacher are similar only when the input examples belong to the same class. In order to model these constraints properly, we propose to employ triplet loss~\cite{Schroff-CVPR-2015}.
Triplet loss has been previously used~\cite{Schroff-CVPR-2015} to obtain close embeddings for objects belonging to the same class and distant embeddings for objects belonging to different classes, while ignoring other factors of variation such as pose, illumination and rotation. We hereby explain in detail how triplet loss can be adapted and used to distill knowledge from a teacher neural network to a student neural network.

In the following, we use the prime symbol to distinguish between fully-visible and occluded images. For a fully-visible input image $x$ and an occluded input image $x'$, let $E_T(x)$ and $E_S(x')$ denote the embedding vectors (face embeddings or embeddings) produced by the teacher network $T$ and the student network $S$, respectively. In order to distill knowledge using triplet loss, we need triplets of input images of the form $(a', p, n')$, where $a'$ is an occluded image from a class $k$, called anchor example, $p$ is a fully-visible image from the same class $k$ as the anchor, called positive example, and $n'$ is another occluded image from a class $j \neq k$, called negative example. During training, our goal is to reduce the distance between the anchor embedding $E_S(a')$ and the positive embedding $E_T(p)$, so that it becomes smaller than the distance between the anchor embedding $E_S(a')$ and the negative embedding $E_S(n')$. We thus optimize the following triplet loss function:
\begin{equation}\label{eq_loss_triplet}
\begin{split}
\mathcal{L}_{triplet}(\theta_S) = \sum_{i=1}^{m} \Big[ &\Delta(E_S(a'_i), E_T(p_i)) - \\
&- \Delta(E_S(a_i'), E_S(n'_i)) + \alpha \Big]_+,
\end{split}
\end{equation}
where $\theta_S$ are the weights of the student network $S$, $m$ is the number of training examples, $\Delta$ is the Euclidean distance, $[\cdot]_+$ is equal to $max(0, \cdot)$ and $\alpha$ is a margin that is enforced between the positive pair $(a'_i,p_i)$ and the negative pair $(a'_i,n'_i)$. In the experiments, we set the margin $\alpha$ to $0.1$.

In a similar fashion to the standard teacher-student loss from Equation~\eqref{eq_loss_KD}, we also want to make sure that the student network predicts the labels correctly. Therefore, our final loss function is:
\begin{equation}\label{eq_loss_KD_triplet}
\mathcal{L}_{KD''} (\theta_S) = \mathcal{L}(y, N_S) + \lambda \mathcal{L}_{triplet}(\theta_S),
\end{equation}
where $\theta_S$ are the weights of the student network $S$, $\mathcal{L}(y, N_S)$ is the cross-entropy with respect to the correct labels $y$, $\mathcal{L}_{triplet}$ is the triplet loss defined in Equation~\eqref{eq_loss_triplet} and $\lambda$ is a parameter that controls the importance of the triplet loss with respect to the cross-entropy loss. In the experiments, we set $\lambda$ to $0.5$.

For an efficient training process, we generate the triplets as follows. Each occluded image from the training set is selected as the anchor $a'$. For each anchor, we sample a random subset of fully-visible images belonging to the same class as the anchor, we compute the distances between the anchor and each image in the resulting subset and we select the image that is farthest from the anchor as the positive example $p$. In a similar fashion, we sample a random subset of occluded images belonging to different classes than the anchor, we compute the distances between the anchor and each image in the resulting subset and we select the image that is closest to the anchor as the negative example $n'$. In both cases, the random subsets represent only one tenth of the full training set, reducing the processing time by an order of magnitude. The random subsets are sampled once per epoch. The effect of training using the defined triplet loss is illustrated in Figure~\ref{fig_triplet_loss}. We note that only the weights $\theta_S$ of the student model are updated during training, hence the embedding $E_T(p_i)$ cannot be modified.

\section{Experiments}
\label{sec_experiments}
%\vspace{-0.1cm}
\subsection{Data Sets}
%\vspace{-0.1cm}

\noindent
{\bf FER+.}
The FER+ data set~\cite{Barsoum-ICMI-2016} is a newer version of the FER 2013 data set~\cite{Goodfellow-ICONIP-2013} which contains images with wrong labels as well as images not containing faces. Barsoum et al.~\cite{Barsoum-ICMI-2016} cleaned up the FER 2013 data set by relabeling images and by removing those without faces. In the relabeling process, Barsoum et al.~\cite{Barsoum-ICMI-2016} added a new class of emotion, \textit{contempt}, while also keeping the other $7$ classes from FER 2013: \textit{anger}, \textit{disgust}, \textit{fear}, \textit{happiness}, \textit{neutral}, \textit{sadness}, \textit{surprise}. The FER+ data set is composed of $25045$ training images, $3191$ validation images and $3137$ test images. The size of each image is $48 \times 48$ pixels.

\noindent
{\bf AffectNet.}
The AffectNet~\cite{Mollahosseini-TAC-2019} data set is one of the largest data sets for facial expression recognition, containing $287651$ training images and $4000$ validation images with manual annotations. The images from AffectNet have various sizes. Since the test set is not yet publicly available, methods~\cite{Mollahosseini-TAC-2019,Zeng-ECCV-2018} are commonly evaluated on the validation set. The data set contains the same $8$ classes of emotion as FER+. With $500$ images per class in the validation set, the class distribution is balanced. In the same time, the training data is unbalanced. As proposed by Mollahosseini et al.~\cite{Mollahosseini-TAC-2019}, we down-sample the training set for classes with more than $15000$ images. This leaves us with a training set of $88021$ images. 

\subsection{Data Preprocessing}

In order to train and evaluate neural models in the VR setting, we replace the values of occluded pixels with zero. For the setting proposed by Georgescu et al.~\cite{Georgescu-ICONIP-2019}, in which facial expressions are recognized from the lower-half of the face, we occlude the entire upper-half of the FER+ and AffectNet images. For the setting proposed by Hickson et al.~\cite{Hickson-WACV-2019}, in which facial expressions are recognized from the eyes region, we occlude the entire lower-half of the FER+ and AffectNet images. Our own approaches are trained and tested in the former setting, in which the upper-half is occluded. All images are resized to $224 \times 224$ pixels, irrespective of the data set, in order to be given as input to VGG-f or VGG-face.

%\vspace{-0.2cm}
\subsection{Implementation Details}
%\vspace{-0.1cm}

All neural models are trained with stochastic gradient descent with momentum. We set the momentum rate to $0.9$. The VGG-face model is trained on mini-batches of $64$ images, while the VGG-f model is trained on mini-batches of $512$ images. We use the same mini-batch sizes in all training stages.

\noindent
{\bf Preliminary training of teachers and students.}
For the preliminary fine-tuning of the teacher and the student models, we use the MatConvNet~\cite{matconvnet} library. The teacher VGG-face is fine-tuned on facial expression recognition from fully-visible faces for a total of $50$ epochs. The teacher VGG-f is fine-tuned for $800$ epochs. We note that VGG-f requires more training time to converge because it is pre-trained on a more distant task, object recognition, unlike VGG-face, which is pre-trained on face recognition. The student VGG-face is fine-tuned on facial expression recognition from occluded faces for $40$ epochs. Similarly, the student VGG-f is fine-tuned for $80$ epochs. Further details about training VGG-face and VGG-f on fully-visible or occluded faces are provided in~\cite{Georgescu-ICONIP-2019}.

\noindent
{\bf Standard teacher-student training.} 
For the standard teacher-student strategy, the student VGG-face is trained for $50$ epochs starting with a learning rate of $10^{-4}$, decreasing it when the validation error does not improve for $10$ consecutive epochs. By the end of the training process, the learning rate for the student VGG-face drops to $10^{-5}$. In a similar manner, the student VGG-f is trained for $200$ epochs starting with a learning rate of $10^{-3}$, decreasing it when the validation error does not improve for $10$ consecutive epochs. By the end of the training process, the learning rate for the student VGG-f drops to $10^{-4}$.

\noindent
{\bf Teacher-student training with triplet loss.}
In order to apply the teacher-student training based on triplet loss, we switch to TensorFlow \cite{Abadi-OSDI-2016}, exporting the VGG-face and VGG-f models from MatConvNet. We train the student VGG-face for $10$ epochs using a learning rate of $10^{-6}$. In a similar fashion, we train the student VGG-f for $10$ epochs using a learning rate of $10^{-5}$.

\noindent
{\bf Combining distilled embeddings.}
After training the student models using the two teacher-student strategies independently, we combine the corresponding embeddings into a single embedding vector through concatenation. The concatenated embeddings are provided as input to a linear Support Vector Machines (SVM) model~\cite{cortes-vapnik-ml-1995}. We set the regularization paramaeter of the SVM to $C=1$. We use the SVM implementation from Scikit-learn~\cite{Pedregosa-JMLR-2011}.

\subsection{Baselines}

As baselines, we consider two state-of-the-art methods~\cite{Georgescu-ICONIP-2019,Hickson-WACV-2019} designed for facial expression recognition in the VR setting. The key contribution of these methods resides in the region they use to extract features, the lower-half of the face (mouth region)~\cite{Georgescu-ICONIP-2019} or the upper-half of the face (eyes region)~\cite{Hickson-WACV-2019}. In order to conduct a fair comparison, we use the same neural architectures for both baselines and our approach.

As reference, we include some results on FER+ and AffectNet from the recent literature~\cite{Barsoum-ICMI-2016,Mollahosseini-TAC-2019,Guo-Sensors-2020,Ionescu-WREPL-2013,Kollias-BMVC-2019}. We note that these state-of-the-art methods~\cite{Barsoum-ICMI-2016,Mollahosseini-TAC-2019,Guo-Sensors-2020,Ionescu-WREPL-2013,Kollias-BMVC-2019} are trained and tested on fully-visible faces. Hence, the comparison to our approach or other approaches applied on occluded faces~\cite{Georgescu-ICONIP-2019,Hickson-WACV-2019} is unfair, but we included it as a relevant indicator of the upper bound for the models applied on occluded faces.

%\vspace{-0.2cm}
\subsection{Results}
%\vspace{-0.1cm}

\begin{table}[!t]
%\vspace{-0.1cm}
\caption{Accuracy rates of various models on AffectNet~\cite{Mollahosseini-TAC-2019} and FER+~\cite{Barsoum-ICMI-2016}, for fully-visible faces (denoted by {\Circle}), lower-half-visible faces (denoted by {\protect\rotatebox[origin=c]{90}{\RIGHTcircle}}) and upper-half-visible faces (denoted by {\protect\rotatebox[origin=c]{90}{\LEFTcircle}}). The VGG-f and VGG-face models based on our teacher-student (T-S) training strategies are compared with state-of-the-art methods~\cite{Barsoum-ICMI-2016,Mollahosseini-TAC-2019,Guo-Sensors-2020,Ionescu-WREPL-2013,Kollias-BMVC-2019} tested on fully-visible faces and with methods~\cite{Georgescu-ICONIP-2019,Hickson-WACV-2019} designed for the VR setting (tested on occluded faces). The test results of our student networks that are significantly better than the stronger baseline~\cite{Georgescu-ICONIP-2019}, according to a paired McNemar's test \cite{Dietterich-NC-1998}, are marked with $\dagger$ for a significance level of $0.05$.}\label{tab_results}
\setlength\tabcolsep{2.5pt}
\begin{center}
\begin{tabular}{|l|c|c|c|}
\hline
\bf Model       & \bf Test faces     & \bf AffectNet     & \bf FER+\\
\hline % 
\hline
Bag-of-visual-words~\cite{Ionescu-WREPL-2013}   & {\Circle} & $48.30\%$ & $80.65\%$\\
VGG-13~\cite{Barsoum-ICMI-2016}                 & {\Circle} & -        & $84.99\%$\\
AlexNet~\cite{Mollahosseini-TAC-2019}           & {\Circle} & $58.00\%$ & - \\   
MT-VGG~\cite{Kollias-BMVC-2019}                 & {\Circle} & $54.00\%$         & - \\
VGG-12~\cite{Guo-Sensors-2020}                  & {\Circle}  & $58.50\%$         & - \\
\hline
Teacher VGG-f                                   & {\Circle}    & $57.37\%$         & $85.05\%$\\
Teacher VGG-face                                & {\Circle}    & $59.03\%$         & $84.79\%$\\ 
\hline
Teacher VGG-f           & {\rotatebox[origin=c]{90}{\RIGHTcircle}}   & $41.58\%$         & $70.00\%$\\
Teacher VGG-face        & {\rotatebox[origin=c]{90}{\RIGHTcircle}}   & $37.70\%$         & $68.89\%$\\
\hline
Teacher VGG-f           & {\rotatebox[origin=c]{90}{\LEFTcircle}}   & $26.85\%$         & $40.07\%$\\
Teacher VGG-face        & {\rotatebox[origin=c]{90}{\LEFTcircle}}   & $31.23\%$         & $48.29\%$\\
\hline
VGG-f~\cite{Georgescu-ICONIP-2019}              & {\rotatebox[origin=c]{90}{\RIGHTcircle}}  & $47.58\%$         & $78.23\%$\\
VGG-face~\cite{Georgescu-ICONIP-2019}           & {\rotatebox[origin=c]{90}{\RIGHTcircle}}  & $49.23\%$         & $81.28\%$\\
\hline
VGG-f~\cite{Hickson-WACV-2019}                  & {\rotatebox[origin=c]{90}{\LEFTcircle}}  & $42.45\%$         & $66.18\%$\\
VGG-face~\cite{Hickson-WACV-2019}               & {\rotatebox[origin=c]{90}{\LEFTcircle}}  & $43.18\%$         & $70.19\%$\\
\hline
VGG-f (standard T-S)                         & {\rotatebox[origin=c]{90}{\RIGHTcircle}}  & $48.75\%^\dagger$         & $80.17\%^\dagger$\\
VGG-face (standard T-S)                      & {\rotatebox[origin=c]{90}{\RIGHTcircle}}  & $49.75\%$         & $82.37\%$\\
\hline
VGG-f (triplet loss T-S)                            & {\rotatebox[origin=c]{90}{\RIGHTcircle}}  & $48.13\%$         & $80.05\%^\dagger$\\
VGG-face (triplet loss T-S)                         & {\rotatebox[origin=c]{90}{\RIGHTcircle}}  & $49.71\%$         & $82.57\%$\\
\hline
VGG-f (triplet loss + standard T-S)          & {\rotatebox[origin=c]{90}{\RIGHTcircle}}  & $48.70\%^\dagger$         & $81.09\%^\dagger$\\
VGG-face (triplet loss + standard T-S)         & {\rotatebox[origin=c]{90}{\RIGHTcircle}}  &  $50.09\%^\dagger$        & $82.75\%^\dagger$\\
\hline
\end{tabular}
\end{center}
%\vspace{-0.2cm}
%\vspace{-0.3cm}
\end{table}

\begin{figure*}[!t]
\centering
\includegraphics[width=0.65\linewidth]{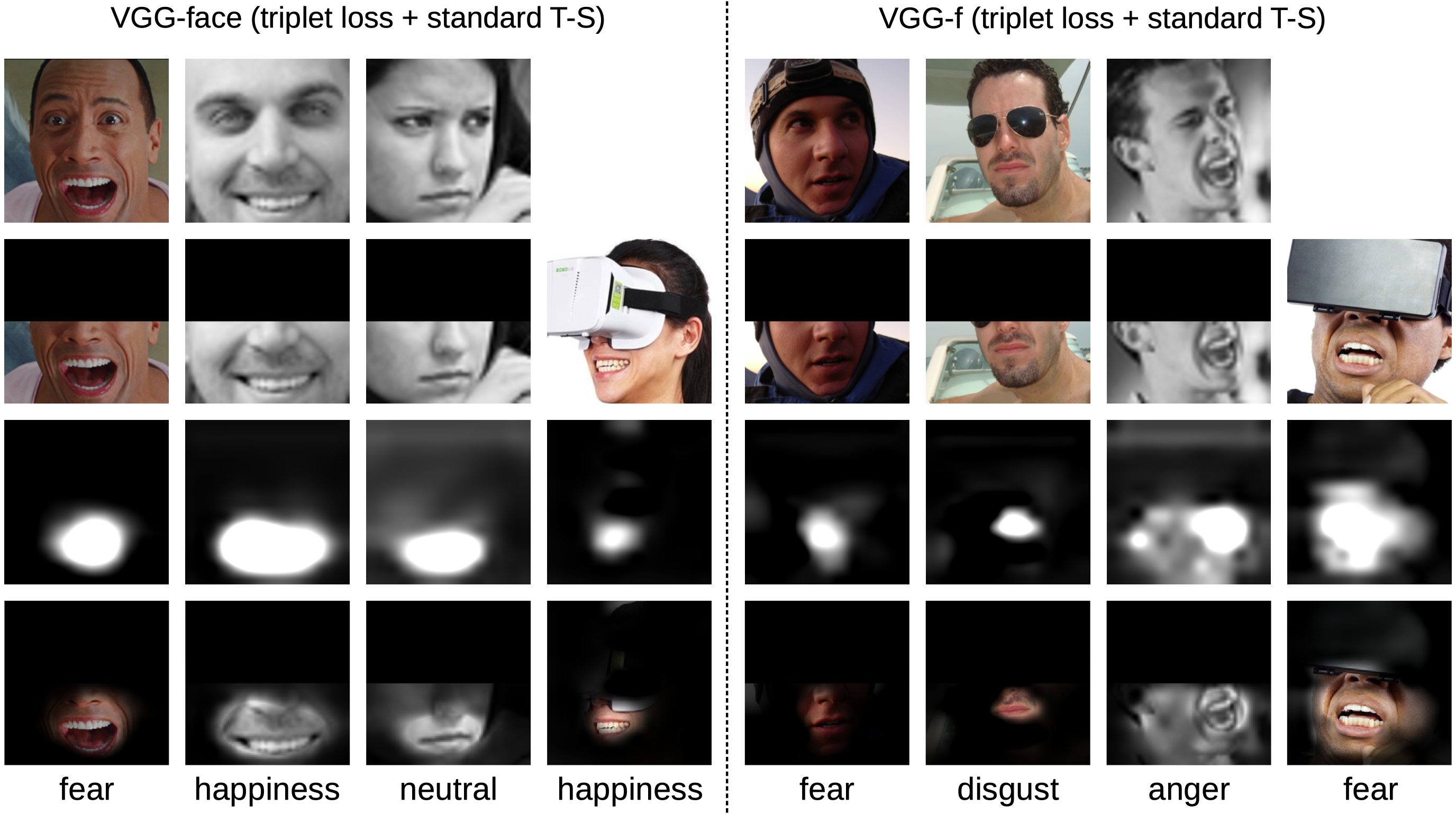}
%\vspace{-0.3cm}
\caption{Fully-visible images ({\Circle}) on top row, lower-half-visible faces ({\protect\rotatebox[origin=c]{90}{\RIGHTcircle}}) on second row, Grad-CAM~\cite{Selvaraju-ICCV-2017} explanation masks on third row and lower-half-visible faces with superimposed Grad-CAM masks on bottom row. The predicted labels provided by the VGG-face (left-hand side) or the VGG-f (right-hand side) models based on the combined teacher-student training strategies are also provided at the bottom. The first three examples from each side are selected from AffectNet~\cite{Mollahosseini-TAC-2019} and FER+~\cite{Barsoum-ICMI-2016}. The fourth example from each side is a person wearing an actual VR headset. Best viewed in color.}\label{fig_visual_explain}
%\vspace{-0.3cm}
\end{figure*}

In Table~\ref{tab_results}, we present the empirical results obtained on AffectNet~\cite{Mollahosseini-TAC-2019} and FER+~\cite{Barsoum-ICMI-2016} by the VGG-f and VGG-face models based on our teacher-student training strategies in comparison with the results obtained by the state-of-the-art methods~\cite{Barsoum-ICMI-2016,Mollahosseini-TAC-2019,Guo-Sensors-2020,Ionescu-WREPL-2013,Kollias-BMVC-2019} tested on fully-visible faces and by the methods~\cite{Georgescu-ICONIP-2019,Hickson-WACV-2019} designed for the VR setting (tested on occluded faces). First of all, we note that it is natural for the state-of-the-art methods~\cite{Barsoum-ICMI-2016,Mollahosseini-TAC-2019,Guo-Sensors-2020,Ionescu-WREPL-2013,Kollias-BMVC-2019} to achieve better accuracy rates than our approach or the other approaches applied on occluded faces~\cite{Georgescu-ICONIP-2019,Hickson-WACV-2019}. An exceptional case is represented by the VGG-face model of Georgescu et al.~\cite{Georgescu-ICONIP-2019} and our student VGG-face models, which surpass the bag-of-visual-words model~\cite{Ionescu-WREPL-2013} on both data sets.

With respect to the baselines~\cite{Georgescu-ICONIP-2019,Hickson-WACV-2019} designed for the VR setting, all out teacher-student training strategies provide superior results. We observe that the accuracy rates of Hickson et al.~\cite{Hickson-WACV-2019} are considerably lower than the accuracy rates of Georgescu et al.~\cite{Georgescu-ICONIP-2019} (differences are between $5\%$ and $12\%$), although the neural models have identical architectures. We hypothesize that this difference is caused by the fact that it is significantly harder to recognize facial expressions from the eyes region (denoted by {\rotatebox[origin=c]{90}{\LEFTcircle}}) than from the mouth region (denoted by {\rotatebox[origin=c]{90}{\RIGHTcircle}}). To test this hypothesis, we evaluate the teacher VGG-f and the teacher VGG-face on upper-half-visible and lower-half-visible faces. We observe even larger differences between the results on upper-half-visible faces (accuracy rates are between $26\%$ and $49\%$) and the results on lower-half-visible faces (accuracy rates are between $37\%$ and $70\%$), confirming our hypothesis. We also note that the results attained by the teacher models on occluded faces are much lower than the results of the baselines~\cite{Georgescu-ICONIP-2019,Hickson-WACV-2019} designed for the VR setting, although the teacher models attain results on par with the state-of-the-art methods~\cite{Barsoum-ICMI-2016,Mollahosseini-TAC-2019,Guo-Sensors-2020,Kollias-BMVC-2019} when testing is performed on fully-visible faces. This indicates that CNN models trained on fully-visible faces are not particularly suitable to handle severe facial occlusions, justifying the need for training on occluded faces.

The results presented in the last six rows of Table~\ref{tab_results} indicate that the teacher-student learning strategies provide very good results on lower-half-visible faces, surpassing the other methods~\cite{Georgescu-ICONIP-2019,Hickson-WACV-2019} evaluated on occluded faces. We believe that the accuracy gains are due to the teacher neural networks that are trained on fully-visible images, which bring additional (privileged) information from the (unseen) upper half of the training faces. Our teacher-student training strategy based on triplet loss provides results that are comparable to the standard teacher-student training strategy. Nevertheless, we achieve additional performance gains when the two teacher-student strategies are combined through embedding concatenation. Our final models based on the concatenated distilled embeddings attain results that are close to the state-of-the-art methods~\cite{Barsoum-ICMI-2016,Mollahosseini-TAC-2019,Guo-Sensors-2020,Kollias-BMVC-2019}. For example, our VGG-face with triplet loss and standard teach-student training yields an accuracy rate of $82.75\%$ on FER+, which is $2.24\%$ under the state-of-the-art VGG-13~\cite{Barsoum-ICMI-2016}. We thus conclude that our models can recognize facial expressions with sufficient reliability, despite being tested on faces that are severely occluded (the entire upper-half is occluded).

We also performed statistical significance testing to compare our VGG-f and VGG-face models based on teacher-student training with the VGG-f and VGG-face models of Georgescu et al.~\cite{Georgescu-ICONIP-2019}. We note that the combined teacher-student strategies provide significant improvements for both models on both data sets, with a significance level of $0.05$.

In order to better understand how our models based on concatenated distilled embeddings make decisions, we used the Grad-CAM~\cite{Selvaraju-ICCV-2017} approach to provide visual explanations for some image examples illustrated in Figure~\ref{fig_visual_explain}. First, we notice that we, as humans, are still able to recognize the facial expressions in the presented examples, even if the upper half of the faces depicted in the second row of Figure~\ref{fig_visual_explain} are occluded. We observe that both neural architectures focus their attention on the lower part of the face, particularly on the mouth region. This indicates that our neural networks can properly handle situations in which people wear VR headsets. Notably, all the predicted labels for the samples presented in Figure~\ref{fig_visual_explain} are correct.

%\vspace{-0.2cm}
\section{Conclusion}
\label{sec_conclusion}
%\vspace{-0.2cm}

In this paper, we proposed to train neural networks for facial expression recognition under strong occlusion, by applying two teacher-student strategies. To our knowledge, this is a novel application of teacher-student training. Another novel contribution is the second teacher-student strategy, which is based on triplet loss. The empirical results indicate that our teacher-student strategies can bring significant performance gains over the baselines~\cite{Georgescu-ICONIP-2019,Hickson-WACV-2019}, particularly when the strategies are combined by concatenating the corresponding distilled embeddings. Notably, on the FER+ data set, our VGG-face based on concatenated distilled embeddings attains an accuracy rate of $82.75\%$ on lower-half-visible faces, which is only $2.24\%$ below the accuracy rate of the state-of-the-art VGG-13~\cite{Barsoum-ICMI-2016} on fully-visible faces. In conclusion, we consider that our results are high enough to deploy our neural models in production.

\section*{Acknowledgment}
Research supported by Novustech Services through Project 115788 (Innovative Platform based on Virtual and Augmented Reality for Phobia Treatment) funded under the POC-46-2-2 by the European Union through FEDR.



%\bibliographystyle{IEEEtran}
\bibliographystyle{IEEEtran}
\bibliography{references}

\end{document}